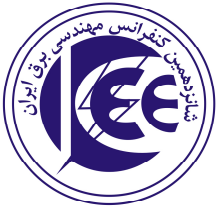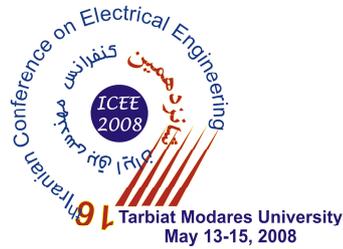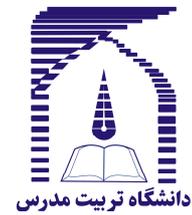

# Combined A*-Ants Algorithm: A New Multi-Parameter Vehicle Navigation Scheme

**Hojjat Salehinejad, Hossein Nezamabadi-pour, Saeid Saryazdi and Fereydoun Farrahi-Moghaddam**
Department of Electrical Engineering, Shahid Bahonar University of Kerman, Kerman, Iran
h.salehi@mail.uk.ac.ir, nezam@mail.uk.ac.ir, saryazdi@mail.uk.ac.ir, ffarrahi@mail.uk.ac.ir

***Abstract:*** *In this paper a multi-parameter A*(A-star)-ants based algorithm is proposed in order to find the best optimized multi-parameter path between two desired points in regions. This algorithm recognizes paths, according to user desired parameters using electronic maps. The proposed algorithm is a combination of A* and ants algorithm in which the proposed A* algorithm is the prologue to the suggested ant based algorithm .In fact, this A* algorithm invigorates some paths pheromones in ants algorithm. As one of implementations of this method, this algorithm was applied on a part of Kerman city, Iran as a multi-parameter vehicle navigator. It finds the best optimized multi-parameter direction between two desired junctions based on city traveler parameters. Comparison results between the proposed method and ants algorithm demonstrates efficiency and lower cost function results of the proposed method versus ants algorithm.*

## 1. Introduction

The development of the vehicle navigation systems started since early 1990. The vehicle navigation system is an information system in substance. Therefore it is necessary of the frame structure, organization and management of electronic map data, as all powerful system functions must be run based on it [2]. Electronic map data should include information such as medical treatment, entertainment, number of traffic lights, traffic flow and accidents occurred history [1,2], useful for finding paths.

Pathfinding plays an important role in many modern games. It is usually a state-space search applied to a two-, or three-dimensional map where each state describes a position on the map [3]. One of the most game programmers popular algorithms is A* (A-star) algorithm. This is due to the fact that it is often the algorithm of choice. It is also popular because of easy implementation, efficiency and huge body of experience [6].

On the other side, ants algorithm has plenty of applications in pathfinding and network problems such as transportation networks [10] and communication networks [9]. Although real ants are blind, they are capable of finding the shortest path from food source to their nest by exploiting information of a liquid substance called pheromone, which they release on the route transiting in.

The idea of employing the foraging behavior of ants as a method of stochastic combinatorial optimization was initially introduced by Dorigo in his PhD thesis [4].

In this paper, an A*-ants based algorithm is proposed in order to find the best optimized multi-parameter direction between two desired points using electronic maps. As the problem is complex, the proposed A* algorithm is the prologue to the proposed ants algorithm. In other words, A* algorithm is run before ants algorithm and updates (increases) pheromones of its resulted paths in ants algorithm. This algorithm is applied on a part of Kerman city, Iran, for multi-parameter desires. Comparison results between the proposed method and the ants algorithm demonstrates efficiency and lower cost results of the proposed method versus ants algorithm

This paper is organized as follows. The next section reviews the basic principles of the A* and ants algorithm. Section 3 represents the details of the proposed method and experimental results are demonstrated in section 4. Finally, the paper in concluded in section 5.

## 2. A review on A* algorithm and Ant colony system

### 2.1 A* algorithm

A* was the result of two decades of research into state-space algorithms [3]. A simple form of A* algorithm is described briefly in this section. The search area is a two-dimensional square grid area with an obstacle between two desired points *X* and *Y* as in Figure (1).

Assume an agent is looking for shortest direction from point *X* to point *Y*. The steps are as follow.

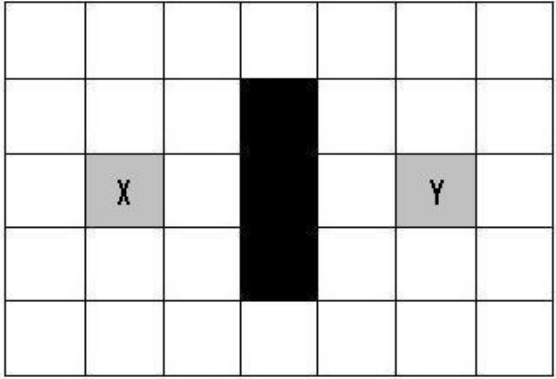

Fig.1 A simple two-dimensional square grid area

a) The start point, *X*, is labeled as *parent square* and is added to the *open* list with walkable squares adjacent to the starting point.
b) The point *X* is dropped from *open* list and added to the *closed* list.
c) A square is chosen from the *open* list by means of the Equation (1).

$$f = g + h \quad (1)$$

Where $g$ is the movement score from *parent* to its adjacent walkable squares; $h$ is the heuristic movement score from the origin to the destination using *Manhattan* method. In these procedures, scores are calculated by considering each vertical or horizontal movement score 10-unit, and each diagonal movement score 14-unit.

d) The lowest $f$ score square is dropped from the *open* list and is added to the *closed* list. The mentioned procedure is repeated until arrival to the destination. The squares in the *closed* list are the direction.

### 2.2 Ant colony system

Ant colony optimization (ACO) is a class of algorithms, whose first member, called Ant System (AS), was initially proposed by Colorni, Dorigo and Maniezzo [8]. The AS strategy developed by Dorigo attempts to simulate the behavior of real ants with the addition of several artificial characteristics: visibility, memory and discrete time to solve many complex problems successfully such as the traveling salesman problem (TSP) [5], vehicle routing problem (VRP) [11] and best path planning [7]. Even though many changes have applied to ACO algorithms during past few years, but their fundamental ant behavioral mechanism which is positive feedback process demonstrated by a colony of ants, is still the same.

Different steps of an ant colony system algorithm (ACSA) are the followings.
- Problem graph representation: As problems which could be solved by ACSA are often discrete, they could be represented by a graph with N nodes and R routes, $G = \langle N, R \rangle$.
- Initializing ant distribution: A number of ants are placed on the origin nodes.
- Node transition rule: The node transition rule specifies ants movement from node to node which is probabilistic. The probability of displacing ant $k$ from node $i$ to node $j$ is calculated by Equation (2).

$$p_{ij}^k = \begin{cases} \dfrac{(\tau_{ij})^\alpha (\eta_{ij})^\beta}{\sum_{h \notin tabu_k}(\tau_{ih})^\alpha (\eta_{ih})^\beta} & if \ j \notin tabu_k \\ 0 & otherwise \end{cases} \quad (2)$$

Where $\tau_{ij}$ and $\eta_{ij}$ are respectively the intensity of pheromone and the visibility of direct rout from $i$ to $j$. $\alpha$ and $\beta$ are parameters which control the relative importance of $\tau_{ij}$ and $\eta_{ij}$ respectively. $tabu_k$ is set of unavailable routes for ant $k$.

- Update Global Trail: A cycle of ACO algorithm is completed when every ant has assembled a solution. At the end of each cycle, the intensity of pheromone is updated by a pheromone trail updating rule as in Equation (3).

$$\tau_{ij}(new) = (1-\rho)\tau_{ij}(old) + \sum_{k=1}^{m} \Delta\tau_{ij}^k \quad (3)$$

Where $\rho$ is a constant parameter $(0 < \rho < 1)$ named pheromone evaporation and $\Delta\tau_{ij}^k$ is the amount of pheromone laid on route $(i,j)$ by the $k^{th}$ ant and could be given by Equation (4).

$$\Delta \tau_{ij}^k = \begin{cases} \dfrac{Q}{f_k} & \text{if route (i, j) be traversed by the } k^{th} \text{ ant (at the current cycle)} \\ 0 & \text{otherwise} \end{cases} \quad (4)$$

Where $f_k$ is the cost value of the solution found by ant *K*.
- Stopping criterion: The end of the algorithm could be achieved by a predefined number of cycles, or the maximal number of cycles between two improvements of the global best solutions.

## 3. Proposed method

In this paper an A*-ants based algorithm is proposed. This algorithm is a multi-parameter navigator using electronic maps in which finds the best optimized direction between two desired points based on users needs. This algorithm is exclusively applied on a part of Kerman city as a vehicle navigation algorithm. In this method, A* algorithm finds some candidate directions by solving the problem. Afterwards, the results are entered into the ants algorithm. In fact, the A* algorithm influences the ants algorithm by updating (increasing) pheromone of resulted directions.

It was attempted to make a collection of all parameters important for travelers taking journey in cities. These parameters are distance, width, traffic load, road risk, medical treatment, entertainment, number of traffic lights, traffic flow and history of accidents occurred. An algorithm named ant-based vehicle navigation (AVN) algorithm was proposed in [1] which finds the best optimized directions based on ants algorithms. However, the algorithm proposed here is an upgrade version of AVN algorithm. As it was mentioned before, this algorithm consists of two algorithms itself. These algorithms are A* algorithm and ants algorithm. A* algorithm is a prologue of the ants algorithm. It invigorates some produced directions by itself in order to help ants algorithm, recognize best direction with higher reliability and lower cost than ants algorithm. This is done by updating (increasing) pheromone amount of directions founded in the A* algorithm. A pseudocode of this algorithm is presented in Figure (2).

```
Procedure A*-ants algorithm
   Initialize
   Activate A* algorithm
   Update pheromone list
   Active ants algorithm
   Express best optimized direction
End A*-ants algorithm
```

Fig 2.The proposed A*-ants algorithm pseudocode

Updating pheromone amount for resulted direction from intersection $i$ to $j$ ($\tau_{ij}$) in the proposed A* algorithm is done by the Equation (5).

$$\tau_{ij}(new) = \Delta\tau + \tau_{ij}(old) \quad ij \in T \quad (5)$$

Where $\Delta\tau \geq 1$ is the ants algorithm pheromone updating constant parameter adjusted by try and error. The proposed A* algorithm produced directions are recorded in the list $T$.

The proposed A* algorithm is discussed in more details in the following subsections.

### 3.1. Proposed A* algorithm

The pseudocode of the A* algorithm used in the proposed algorithm is presented in Figure (3).

```
Procedure A* algorithm
Add start to open
While open is not empty
   Let n=first nod on open
   Drop n from open & add it to closed
   Add adjacent walkable squares to open
   Compute scores
   Select next square
   Add parent to closed
   If parent=destination
      Search is succeed
   Else if open is empty and destination is not find
      Search is fail
   End
End
```

Fig.3 The proposed A* algorithm pseudocode

As this pseudocode is clear and some of its steps were defined in the last section, brief descriptions of some of its steps are presented as follow.
- Compute scores: In this stage, all adjacent squares to the *n* square, except impossible and those in the *closed* list scores are calculated. The selected square is named *parent*.
- Select next square: If the *g* score for any left square in the *open* list is better than the current square, the parent is changed to this square. Afterwards, *h* and *g* scores are calculated again for the new *parent*.

This algorithm is successful when open list is empty and destination is found.

### 3.2. Proposed ants algorithm

This subsection discusses the employed ants algorithm in details. The pseudocode of this

156
computer

algorithm used in the proposed algorithm is presented in Figure (4).

```
Procedure ants algorithm
  Initialize
  For each loop
    Locate ants
    For each iteration
      For each ant
        If ant be alive
          Construct probability
          Select route
          Update tabu list
        End
      Next ant
    Next iteration
    Value ants
    Update pheromone
  Next loop
  Select best optimized direction
End ants algorithm
```

Fig.4 The proposed ants algorithm pseudocode.

This pseudo code is contained of the following steps.
- Initialize: It consists of algorithm parameters initial values such as number of ants, evaporation coefficient, and average velocity of vehicle.
- Locate ants: Ants are located on the start point in this stage.

Alive ant refers to an ant which has not arrived to the destination yet and is not blocked in an intersection. An ant is blocked in an intersection is a situation where has no opportunity to continue its transition toward the destination as there is no possible route to continue or transiting backward [1].
- Construct probability: In this step, the probability of each possible direct route is calculated based on its cost function for each alive ant. The probability of displacing from intersection $i$ to intersection $j$ for ant $k$ is calculated by Equation (6)

$$p_{ij}^k = \begin{cases} \dfrac{\tau_{ij}^\alpha \prod_{l \in parameters} cost_{ij_l}^{-2\alpha_l}}{\sum_{h \notin tabu_k} \tau_{ih} \prod_{l \in parameters} cost_{ih_l}^{-2\alpha_l}} & j \notin tabu_k \\ 0 & otherwise \end{cases} \quad (6)$$

Where $\tau_{ij}$ is the direct route pheromone intensity from intersection $i$ to $j$. $\alpha$ controls the importance of $\tau_{ij}$ and $tabu_k$ is set of direct blocked routes. Parameters set consists of distance, width, traffic load, road risk, road quality and other useful parameters [1]. Cost of each parameter ($l$) is $cost_{ij_l}^{-2\alpha_l}$ where significance of each $l$ is adjustable by $\alpha_l$. Data is entered into the algorithm by an electronic map [2].
- Select route: A random parameter $q$ with uniform probability in $[0,1]$ is compared with a parameter $Q$ $(0 \leq Q \leq 1)$. The result picks up one of two selection methods for the alive ant to continue its route to next intersection. Selection methods are mentioned in Equation (7).

$$j = \begin{cases} \arg \max \left( p_{ih}^k \right) & q > Q \\ Roulette \ Wheel \left( p_{ih}^k \right) & otherwise \end{cases}, \quad (7)$$

If $q > Q$ alive ant selects the route with highest probability, else Roulette Wheel rule will be selected to choose the next intersection through probabilities.
- Update tabu list: In this step, the route which $k^{th}$ ant has chosen is added to tabu list in order not to be selected again and its probability will not be calculated anymore.

If $k^{th}$ ant arrives to the destination or be blocked in an intersection, it is omitted from alive ant list. In other words, this step kills the blocked or arrived ant in the current iteration.

Value ants: The $tcost$ is summation of cost from intersection $i$ to intersection $j$ for alive ant $k$, since it has started its journey from origin until its arrival to destination. Based on $tcost$s average ($avg.tcost$) of all the ants, each ant will be added to one of the $AWA$ or $PLA$ lists base on the Equation (8).

$$k \in \begin{cases} AWA & tcost < avg.tcost \\ PLA & otherwise \end{cases} \quad (8)$$

Where $avg.tcost$ is the average of all arrived ants $tcost$s. If $tcost$ of alive ant $k$ be more than $avg.tcost$, it will be added to $AWA$ list, otherwise it will be added to $PLA$ list.
- Update pheromone: This stage consists of local and global pheromone updating sections. The pheromone amount of route between intersections $i$ and $j$ is updated by Equation (9) for ant $k$.

$$\begin{cases} \tau_{ij}(new) = \tau_{ij}(old) + \left(\dfrac{av}{tcost_{ij}}\right) & if \ k \in AWA \\ \tau_{ij}(new) = \tau_{ij}(old) \times pv & if \ k \in PLA \end{cases} \quad (9)$$

Where $pv$ $(0 < pv < 1)$ and $av$ $(av > 1)$ are respectively punishment and awarding amounts.
The last step of each completed loop is global pheromone updating using Equation (10).

$$\tau_{ij}(new) = \rho \tau_{ij}(old) \quad (10)$$

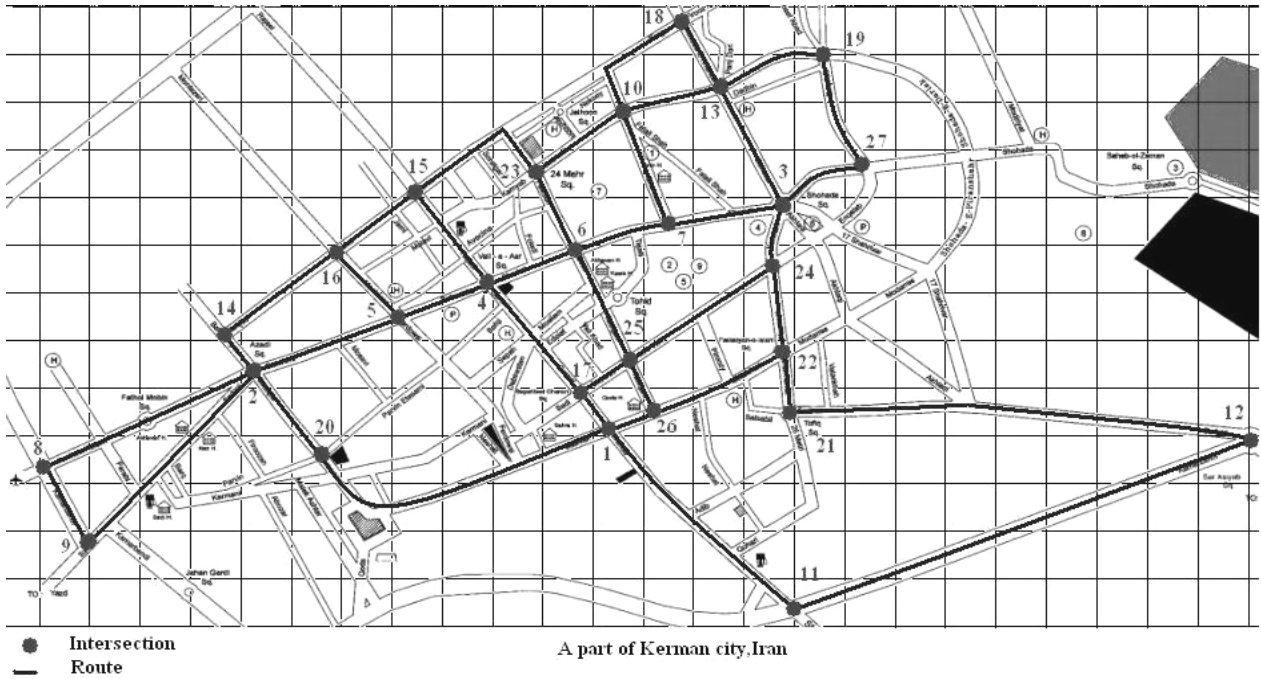

Fig.5 A part of Kerman city, Iran, adapted for experiments.

Where $\rho$ $(0<\rho<1)$ is evaporation coefficient. Select best optimized direction: After *m* loops, the optimized direction with the lowest cost function from origin to destination is the result.
Section 4 presents the simulation results of the proposed algorithm.

## 4. Experimental results

In this section, the proposed algorithm and ants algorithms are compared for finding the best optimized multi-parameter direction in a part of the Kerman city
This procedure is run for two separated parameters importance rates.
The selected part of Kerman city is consisted of 27 intersections as in Figure (5). Suitable parameters used in this algorithm were determined based on trial and error as follows:
$\alpha = 2$, $Q = 0.9$, $pv = 0.9$, $av = 950$ and $\rho = 0.9$.
In the both experiments, the average velocity of vehicle is assumed 40Km/h and start time is 6:00 PM as default. The parameters important rates for both experiments are presented in Table (1) for the origin and destination number 24 and 23 respectively.
The resulted directions from ants algorithm and A*-ants algorithm are as follow respectively.

Table 1

| No. | Parameter | Experience1 | Experience2 |
|---|---|---|---|
| 1 | Distance | 1 | 0.50 |
| 2 | Width | 0.25 | 0.25 |
| 3 | Traffic load | 0.50 | 0.75 |
| 4 | Road risk | 0.25 | 0.75 |
| 5 | Road quality | 0.50 | 0.50 |
| 6 | Traffic lights | 0.25 | 0.25 |

$$24 \rightarrow 22 \rightarrow 26 \rightarrow 25 \rightarrow 6 \rightarrow 23 \qquad (11)$$
$$24 \rightarrow 25 \rightarrow 6 \rightarrow 23 \qquad (12)$$

The cost of directions (11) and (12) are 2486.5 and 812.7 respectively. It should be noticed that the directions from intersection 6 to 7 and 7 to 3 are one-way. Therefore, the proposed algorithm found another direction based on the desired parameter importance rates.
In the second experiment, the resulted directions from ants algorithm and A*-ants algorithm are as follow respectively.

$$24 \rightarrow 3 \rightarrow 13 \rightarrow 10 \rightarrow 23 \qquad (13)$$
$$24 \rightarrow 25 \rightarrow 6 \rightarrow 23 \qquad (14)$$

The cost of directions (13) and (14) are 99.2 and 52.6 respectively. Because the experimental region is simple, the proposed method found similar directions in the both experiments. As it is illustrated in Figures (6,7), the proposed method has less cost average than [1] which is due to the pheromone updating (increasing) of A* algorithm. This figure illustrates efficiency of the proposed method.

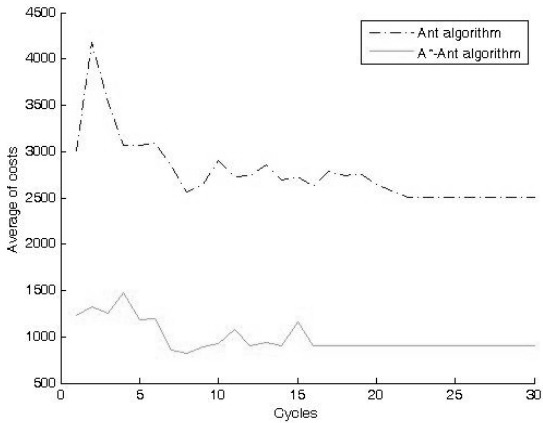

Fig.6 Costs average in the first experiment.

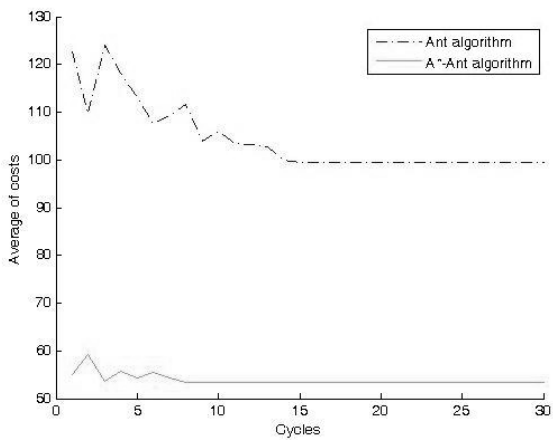

Fig.7 Costs average in the second experiment.

## 5. Conclusion

In this paper a combination of A* and ants algorithm is proposed in order to find the best optimized multi-parameter direction between two desired points. The proposed A*-ants algorithm is a new method where A* algorithm is run before ants algorithm and updates (increases) pheromones of its resulted paths in the ants algorithm. This algorithm was applied on a part of Kerman city, Iran as a vehicle navigation algorithm and the results were encouraging in comparison of ants algorithm.

This method could also be useful for emergency services, taxi drivers, tours and etc. as it is a fast access, low-cost and easy vehicle navigator algorithm.

## Acknowledgements

This paper was supported by Shahid Bahonar University of Kerman Young Researchers Society.